\documentclass{article}

\usepackage{arxiv}

\usepackage[utf8]{inputenc} 
\usepackage[T1]{fontenc}    
\usepackage{hyperref}       
\usepackage{url}            
\usepackage{booktabs}       
\usepackage{amsfonts}       
\usepackage{nicefrac}       
\usepackage{microtype}      
\usepackage{lipsum}		
\usepackage{graphicx}
\usepackage{natbib}
\usepackage{doi}

\usepackage[linesnumbered,ruled,vlined]{algorithm2e}
\usepackage{amsmath,amssymb,amsfonts}
\usepackage{algorithmic}
\usepackage{graphicx}
\usepackage{textcomp}
\usepackage{url}
\usepackage{hyperref}
\usepackage[export]{adjustbox}
\usepackage{gensymb}
\usepackage{subcaption}
\usepackage{booktabs}

\usepackage{balance}
\usepackage{mwe}
\usepackage{float}
\usepackage{listings}
\usepackage{enumitem}
\usepackage{textcomp}
\usepackage{caption}
\usepackage[utf8]{inputenc}

\def\BibTeX{{\rm B\kern-.05em{\sc i\kern-.025em b}\kern-.08em
    T\kern-.1667em\lower.7ex\hbox{E}\kern-.125emX}}

\title{SHEDAD: SNN-Enhanced District Heating Anomaly Detection for Urban Substations}

\author{Jonne van Dreven \\
        Department of Computer Science \\
Blekinge Institute of Technology, Karlskrona, Sweden\\
        \texttt{jonne.van.dreven@bth.se \thanks{Corresponding author.}}
         \And
         Abbas Cheddad \\
        Department of Computer Science \\
Blekinge Institute of Technology, Karlskrona, Sweden\\
        \texttt{abbas.cheddad@bth.se }
         \And
Sadi Alawadi\\
        Department of Computer Science \\
Blekinge Institute of Technology, Karlskrona, Sweden\\
        \texttt{sadi.alawadi@bth.se}
        \And
        Ahmad Nauman Ghazi \\
        Department of Software Engineering\\
Blekinge Institute of Technology, Karlskrona, Sweden\\
        \texttt{nauman.ghazi@bth.se}
         \And
       Jad Al Koussa\\
        Unit Energy Technology, Flemish Institute for Technological Research (VITO), Mol, Belgium\\
        EnergyVille, Genk, Belgium\\
         \And
         Dirk Vanhoudt \\
         Unit Energy Technology, Flemish Institute for Technological Research (VITO), Mol, Belgium\\
        EnergyVille, Genk, Belgium\\
}

\renewcommand{\shorttitle}{SHEDAD: SNN-Enhanced District Heating Anomaly Detection for Urban Substations}

\hypersetup{
pdftitle={A template for the arxiv style},
pdfsubject={q-bio.NC, q-bio.QM},
pdfauthor={David S.~Hippocampus, Elias D.~Striatum},
pdfkeywords={First keyword, Second keyword, More},
}

\begin{document}
\maketitle

\begin{abstract}
District Heating (DH) systems are essential for energy-efficient urban heating. However, despite the advancements in automated fault detection and diagnosis (FDD), DH still faces challenges in operational faults that impact efficiency. This study introduces the Shared Nearest Neighbor Enhanced District Heating Anomaly Detection (SHEDAD) approach, designed to approximate the DH network topology and allow for local anomaly detection without disclosing sensitive information, such as substation locations. The approach leverages a multi-adaptive k-Nearest Neighbor (k-NN) graph to improve the initial neighborhood creation. Moreover, it introduces a merging technique that reduces noise and eliminates trivial edges. We use the Median Absolute Deviation (MAD) and modified z-scores to flag anomalous substations. The results reveal that SHEDAD outperforms traditional clustering methods, achieving significantly lower intra-cluster variance and distance. Additionally, SHEDAD effectively isolates and identifies two distinct categories of anomalies: supply temperatures and substation performance. We identified 30 anomalous substations and reached a sensitivity of approximately 65\% and specificity of approximately 97\%. By focusing on this subset of poor-performing substations in the network, SHEDAD enables more targeted and effective maintenance interventions, which can reduce energy usage while optimizing network performance.
\end{abstract}


\maketitle

\section{Introduction}
\label{sec:introduction}
The high energy consumed by cities worldwide for heating purposes significantly impacts climate change, the surrounding environment and the economy. Essential actions and solutions must be considered to face this challenge. For instance, district Heating (DH) systems offer a solution to meet urban heating demands while leveraging renewable energies efficiently. DH systems generate heat centrally and then distribute it via a network of insulated pipes, making them one of the most sustainable methods for providing heat in densely populated areas~\cite{Werner_2017}. The International Energy Agency (IEA) expects that by 2030, approximately 350 million buildings will be connected to DH systems, fulfilling 20\% of global space heating requirements~\cite{IEA_2022}. Despite their benefits, unlocking the immense possibilities of DH systems will require carefully addressing the substantial challenges that presently constrain their capabilities.

One major challenge is the presence of operational faults in DH substations, which negatively impacts the substation itself and the entire network. Studies indicate that a substantial percentage (approximately 43 and 75\% of the substations) of DH substations can operate sub-optimal due to faults~\cite{Mansson_2019, Gadd_Werner_2015}. The lack of automated Fault Detection and Diagnosis (FDD) methods aggravates this issue, allowing faults to persist for extended periods, significantly impacting energy use. Fortunately, with the introduction of automatic heat meters, DH networks become better equipped for data-driven solutions and, with the proper detection, could help reduce the DH customers'
energy usage on average by 14\%~\cite{Leiria_2023}. Although, current data collection is not standardized and is primarily focused on billing rather than FDD. Commonly, DH datasets comprise primary side hourly measurements, including features such as supply temperature, return temperature, and flow rate, alongside derived metrics such as energy consumption. Most often, secondary side information (supply/return temperatures, flow, set points, and indoor temperature) and supplementary information, such as layouts of DH networks and geographical substation locations, are confidential, limiting the depth and spatial analysis that can be conducted. Additionally, DH data contains much-introduced noise in the measurements, e.g., supply temperature measurements contain substation behavior due to the sensor placement, as well as unknown confounders influencing measurements, e.g., building occupancy, size, or domestic hot water usages, which makes many conventional analytical methods often fall short. Furthermore, the lack of labeled data~\cite{Neumayer_2023} amplifies the challenges, making it difficult to develop Machine Learning (ML) methods and validate them. Given the complexity and constraints of DH data, there is a need for tailored methods specifically for DH. 

Automatic FDD is commonly done through three steps. The process begins with \textit{fault detection}, where algorithms monitor system parameters to identify deviations. This is followed by \textit{fault diagnosis}, which involves analyzing the data to pinpoint the fault's cause. The final step, \textit{fault correction}, adjusts system settings automatically or prompts manual interventions to remedy the identified issues, thereby ensuring system integrity and operational continuity.

 This study introduces an innovative approach, the \textit{Shared Nearest Neighbor Enhanced District Heating Anomaly Detection} (SHEDAD), designed to approximate the relative topology of DH networks. By utilizing local operational data and identifying similarities, SHEDAD significantly enhances anomaly detection precision.

We use the consumer supply temperature profiles as a distance metric to define relative neighbors of DH substations, thereby allowing for localized anomaly detection (intra-cluster) using modified z-scores. We identify two distinct categories of anomalies: those related to supply temperatures and those concerning substation performance. Crucially, our method avoids the need for sensitive data, such as geographical substation locations and network layouts, effectively addressing privacy and security concerns while enabling comparative assessments of substation performance. We advance the traditional Shared Nearest Neighbor (SNN) clustering method by integrating several enhancements tailored for time series DH data. We focus on improving the initial neighborhood creation through a multi-adaptive \(k\)-NN graph, which helps to isolate anomalous substations while improving connectivity for regular substations. Additionally, we introduce a novel merging technique that significantly reduces noise and eliminates meaningless edges that often obscure traditional graphs, thereby clarifying the graph structure and enhancing the effectiveness of the clustering process.

\section{Related Work}
\label{sec:related-work}

Anomaly detection in DH systems typically involves either a global comparison of all substations against each other across the entire system to identify unusual patterns~\cite{mansson_2018, Mansson_2019} or using regression analysis on individual substations to detect anomalies based on fixed thresholds~\cite{Theusch_2021, Calikus_2018}. Regression analysis can detect sudden variations in substation behavior, i.e., when a fault occurs. However, an anomaly may remain undetected if the model is constructed during an existing fault. Local anomaly detection is frequently overlooked in many settings due to the constraints imposed by the confidentiality of geographical substation locations. However, such localized analysis can be particularly insightful for FDD. Comparing the performance of substations that experience similar supply temperatures offers a fairer assessment than comparing those with differing temperatures. Furthermore, understanding the time delays in supply temperature across the network allows for strategic adjustments in flow rates, which can optimize and potentially reduce these delays. Such optimization ensures more efficient heat distribution and enhances the overall operational efficiency of the DH system.

Clustering, such as \(k\)-means and \(k\)-shape, has been widely explored in DH, for instance, to analyze substation heat consumption patterns~\cite{Gianniou_2018,Tureczek_2019,Hong_Yoon_2022}, outlier detection~\cite{Xue_Zhou_2017, Koussa_2022}, or discover heat load patterns~\cite{Calikus_2019}. While \(k\)-means is effective, it is sensitive to noise and outliers and cannot handle non-globular patterns, limiting its performance in complex datasets, such as in DH. Therefore, SNN~\cite{Jarvis_Patrick_1973} could offer advantages when handling DH data, as it focuses on the density of shared nearest neighbors rather than centroid-based distance. This method effectively reduces the impact of outliers and noise, enhancing the cluster results in complex datasets. Recent advancements in SNN have significantly improved the precision and clarity of identifying cluster boundaries. For instance, in~\cite{wu_effective_2021}, a hierarchical clustering-based method utilizing structural similarities in nearest neighbor graphs (HCNN) is introduced, which outperforms traditional methods like the Density-based Spatial Clustering of Applications with Noise (DBSCAN), the Density Peak Clustering (DPC), and the \(k\)-means algorithms in handling unclear boundaries. Similarly, in~\cite{Sengupta_Das_2022}, a Selective Nearest Neighbors Clustering (SNNC) technique is presented, effectively reducing weak connections and enhancing border and outlier detection, thereby outperforming conventional approaches in complex datasets. Furthermore, in~\cite{Liu_Wang_Yu_2018}, an adaptation of the SNN framework combined with DPC utilizes Jaccard similarity metrics to refine cluster center identification and data point assignment, providing a marked improvement over traditional DPC by incorporating nearest-neighbor information for better cluster allocation and center discovery.

This study addresses the limitations of traditional anomaly detection strategies for DH systems, which typically rely on global comparisons that may overlook local nuances. Our SHEDAD method adopts a localized approach through a relative topology approximation, thereby enhancing the precision of anomaly detection and enabling more targeted maintenance interventions.

\section{Data Acquisition}

The dataset used in this study comprises sensor measurements linked to different substations and geographical data of a DH network located in the southern part of Shandong province, China. 
Specifically, the dataset consists of data recorded every 5 minutes from 248 substations. The data collection spanned one month, from 1st January 2024 to 31st January 2024 (8,928 samples), capturing operational measurements of each substation as described in Table~\ref{tab:dataset_features}, across a broad spectrum of outdoor temperatures ranging from \(-9\degree C\) to \(11\degree C\). This wide range of temperatures allows for detailed analysis of the substations under varying climatic conditions. Due to the sensitive nature of the information, the dataset is confidential and has not been made publicly available.

\begin{table}[h]
\centering
\caption{Delineation of the Features Comprising the Dataset}
\label{tab:dataset_features}
\begin{tabular}{@{}lp{1.5cm}p{4.5cm}@{}}
\toprule
\textbf{Feature}        & \textbf{Type}         & \textbf{Description}       \\
\midrule
Timestamp               & Datetime              & Date and time of measurement                   \\
Supply temp.             & Continuous            & Primary supply temperature               \\
Return temp.             & Continuous            & Primary return temperature               \\
Flow                   & Continuous            & Primary flow rate                                 \\
Outdoor Temp.           & Continuous            & Outdoor temperature                   \\ 
X Location              & Continuous              & X coordinate           \\ 
Y location              & Continuous              & Y coordinate           \\ 
\bottomrule
\end{tabular}
\end{table}

\section{Method}
\label{sec:method}

Our approach, SHEDAD, aims to systematically identify two distinct types of anomalies within DH networks: supply temperature and performance anomalies. 

Initially, we construct a relative network topology, which utilizes supply temperature profiles as a distance metric to approximate spatial relationships and operational similarities between substations. This topology serves as the framework for our anomaly detection processes, enabling us to pinpoint local deviations in supply temperature that may indicate underlying issues. The anomalous substations are separated into singleton clusters, while substations with similar supply temperature characteristics are grouped together.

Secondly, leveraging the established network topology, we implement an anomaly detection strategy focused on substation performance. This involves analyzing each substation's operational performance relative to its immediate topological neighbors. We detail each phase of SHEDAD in the subsequent sections. Below, we describe the processes involved in constructing the network topology in subsection~\ref{sec:method:sub:network-topology} and the specific techniques used for subsequent anomaly detection based on substation performance in subsection~\ref{sec:method:sub:anomaly-detection}.

\subsection{Network Topology}
\label{sec:method:sub:network-topology}

The primary supply temperatures reaching the consumer substations are affected by several aspects that contribute to the heat losses~\cite{Zhao_Shan_2019}. The aspects are:

\begin{figure}[]
\centering
\includegraphics[width=\columnwidth]{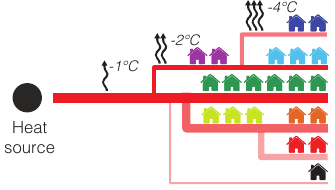}
\caption{Illustration of a DH network with a single heat source and various pipe diameters and flow rates. Substation supply temperatures are affected due to heat loss during transportation. Each color represents a neighboring group of substations with similar supply temperature profiles.}
\label{fig:supply-profiles-illustration}
\end{figure}

\begin{itemize}
    \item Flow rate: The rate at which the heated water or medium is circulated through the network impacts the supply temperature. Higher flow rates can lead to reduced time delays in heat delivery. Flow rate is influenced by a combination of factors related to the heat source, network design, and the consumption patterns of each heat consumer.
    \item Pipe Length and Diameter: Longer distances and variations in pipe diameter affect the time it takes for the heat to reach different parts of the network, which can lead to variations in supply temperature at different substations.
    \item Water Supply Temperature: The initial temperature of the water supplied from the heat source directly influences the supply temperatures throughout the network. Any fluctuations at the source are propagated throughout the system.
\end{itemize}

\begin{figure*}[]
\centering
\includegraphics[width=\textwidth]{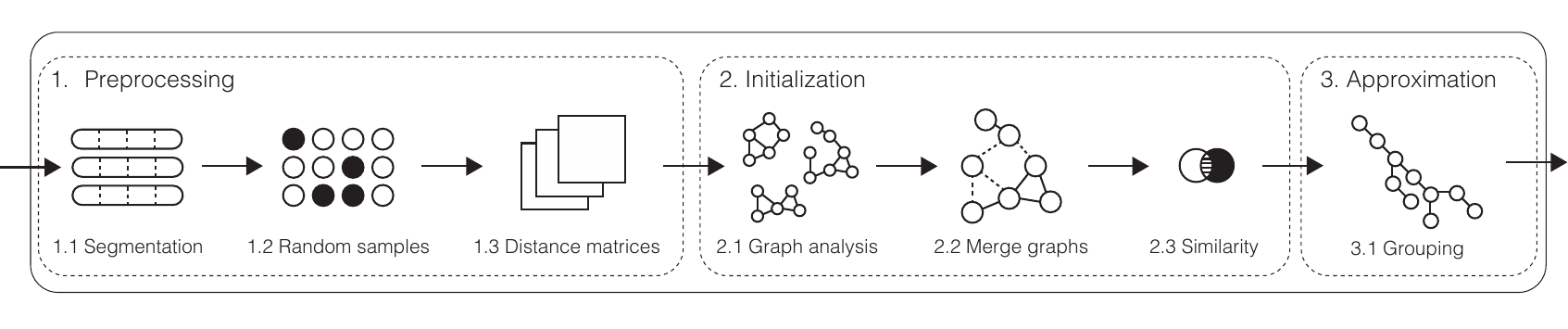}
\caption{Diagram of our proposed method for the relative approximation of a DH network.}
\label{fig:method-illustration}
\end{figure*}

Figure~\ref{fig:supply-profiles-illustration} illustrates a relative network topology with its respective heat losses. Since the combination of these factors becomes unique for each substation, we can use the supply temperature profile to approximate the relative location of a substation to the heat source and its neighboring substations. As shown in Figure~\ref{fig:method-illustration}, the method has three main steps.

\subsubsection{preprocessing}

The preprocessing step prepares the supply temperature profiles from each substation for subsequent analysis. This phase ensures the quality and consistency of the data, which are critical for the reliability of the analysis. At first, we used standard data cleaning procedures, e.g., removing substations with missing values, which resulted in removing 28 substations. 

We start by segmenting the dataset into daily supply profiles. The segmentation allows for studying individual days instead of the entire period, which can help avoid issues that might arise when clustering extensive time series data. For example, significant patterns or anomalies on specific days might get averaged out or diluted if the entire dataset is used.

To ensure our model accurately reflects the diverse operational dynamics of DH networks, it is critical to sample data from a broad spectrum of conditions. For instance, supply temperature variations are influenced by outdoor temperatures, as on colder days, increased flow rates may result in smaller network delays. In contrast, warmer days might see reduced flow and larger network delays. We select \(r\) random days from the segmented data to mitigate selection bias and prevent the model from overfitting. This random selection process helps us capture a representative cross-section of operational scenarios, ensuring our analysis remains robust and generalizable across the different environmental conditions.

For each of the \(r\) randomly selected days, we create a distance matrix between all pairs of substations based on their supply temperature profile. The distances are calculated using Dynamic Time Warping (DTW), with a Sakoe-Chiba warp constraint~\cite{Sakoe_Chiba_1978}. This constraint allows for small time shifts to reflect minor delays in the DH network but prevents overly aggressive realignments that could distort the true relationships. This step is crucial for capturing the dynamic nature of temperature changes over time while ensuring that only meaningful temporal variations are considered to improve the model performance.

\subsubsection{Initialization}

In this step, we aim to construct a representative graph \(G'\) where nodes \(N\) represent substations, and weighted edges \(E\) represent the DTW distance between substations. The primary goal is to optimally balance the number of edges, as an excessive number of edges can obscure vital connections by obscuring the graph. Conversely, too few edges create a sparse graph, potentially overlooking meaningful relationships. Our approach aims to fine-tune the number of edges to capture essential connections effectively, ensuring that the graph remains interpretable and comprehensive without being overly complex or overly simplified.

For each distance matrix \(d_i \in D\), we construct an adaptive k-nearest neighbor (k-NN) graph \(G_i\). The number of neighbors \(k\) for each substation is dynamically adjusted based on its edge weights, allowing more neighbors as long as their weights are below the threshold. Specifically, for substations with high weights, indicating high dissimilarity to other substations, we reduce the associated \(k\) value to minimize substation connectivity, effectively isolating them from the main graph. Conversely, nodes with predominantly low-weight connections, indicating closer proximity, are allowed more neighbors, enhancing their connectivity within the graph. This adaptive approach ensures a nuanced representation of substation relationships, balancing the need to maintain overall network cohesion while isolating outlying substations. The criteria for adjusting \(k_i = k_b + \Delta k_i\), where \( k_b \) is the base number of neighbors, and \( \Delta k_i \) is the adjustment computed as:




\begin{equation}
    \Delta k_i = \begin{cases}
        \left(\frac{\text{low}_i}{k_b} - 1\right) \cdot \frac{k_b}{2}, & \text{if } \text{low}_i \geq k_b, \\
        \left(\frac{\text{high}_i}{k_b} - 1\right) \cdot \frac{k_b}{2}, & \text{if } \text{high}_i \geq k_b,
    \end{cases}
\end{equation}

\noindent where, for node \(i\), \(\text{low}_i\) and \(\text{high}_i\) are the counts of edges with the weight below \(\theta_{\text{min}}\) and above \(\theta_{\text{max}}\), respectively. Then \((i, j) \text{ if } \text{dist}(i, j) \leq \theta_{\text{max}} \text{ and meets the criteria based on } \Delta k_i.\)

The individual \(k\)-NN graphs \(G_i\) for \(i = 1, 2, \ldots, d\) are merged into a single graph \(G'\) using a robust statistical technique namely Fleiss' kappa~\cite{Fleiss_1971}. Fleiss' kappa measures the extent to which the agreement (inclusion) of a certain edge among the individual graphs exceeds what would be expected by chance. This approach ensures that \(G'\) includes only those edges that consistently demonstrate significant agreement across the individual graphs, reducing noise and enhancing the reliability of the merged network effectively.

We calculate a similarity matrix from \(G'\) to quantify the similarity between each pair of substations. Typically, this is done using the Jaccard index~\cite{Jaccard_1912}. However, it solely considers shared neighbors. Therefore, we use a weighted similarity method proposed in~\cite{Liu_Wang_Yu_2018}, which accounts for the number of shared neighbors and the sum of the edge weights connecting them. The weighted similarity between any two nodes \(i\) and \(j\) is defined as:

    
\begin{equation}
    S(i, j) = \frac{|\mathcal{N}(i) \cap \mathcal{N}(j)|^2}{\sum_{k \in SNN(i,j)} (w_{ik} + w_{jk})}
\end{equation}
    
\noindent where \(\mathcal{N}(i) \cap \mathcal{N}(j))\) denote the sets of neighbors for nodes \(i\) and \(j\), respectively, and \(w_{ik}\) and \(w_{jk}\) are the weights of the edges connecting these nodes to their shared neighbor \(k\).

\subsubsection{Approximation}

The final step approximates the relative network topology through analysis of the merged graph obtained earlier. We use Agglomerative Hierarchical Clustering~\cite{sokal1958university} and Ward Linkage~\cite{Ward_1963} to analyze the similarity matrix constructed in the previous step. This method starts with each substation as an independent cluster and iteratively merges them based on the similarity measure provided by the SNN matrix. This method creates a dendrogram, a tree-like diagram illustrating how substations merge during the clustering process. This visualization offers clear, interpretable results revealing the substations' hierarchical relationships. Agglomerative hierarchical clustering is particularly suited for DH networks due to its bottom-up approach, which naturally aligns with how substations are interconnected. By starting with the most similar pairs of substations and progressively building larger clusters, this method mirrors the geographic proximities inherent within the network. This incremental clustering process allows for a nuanced understanding of how local operational characteristics aggregate into broader patterns, providing valuable insights into the complex dynamics of DH systems.

\subsection{Anomaly Detection}
\label{sec:method:sub:anomaly-detection}

The topology approximation forms the essential foundation for our anomaly detection system, making it feasible only with the initial processing in place to enable effective detection. We construct a Minimum Spanning Tree (MST) for each cluster based on pairwise Euclidean distances calculated from the supply temperatures. This MST serves as the framework for our performance comparisons, where each substation in the tree is compared against its \(k\) neighboring substations. Given that a substation may be compared multiple times—up to \(k-1\) times—we normalize the anomaly scores by the number of comparisons each substation undergoes. A normalized score of 1 denotes a significant anomaly, as it has been flagged in all its comparisons. In contrast, a score of 0 indicates average performance relative to its peers, i.e., it has never been flagged as an anomaly.

For each comparison, we calculate the Median Absolute Deviation (MAD) to robustly measure the dispersion within each comparison of \(k + 1\) substations (intra-cluster). Additionally, we use modified z-scores~\cite{Crosby_1994} to identify the outlying substations, i.e., substations with absolute mean z-scores exceeding the threshold of two standard deviations below the median are identified as anomalies. This non-parametric approach provides a robust defense against outliers and ensures a more precise measurement of dispersion compared to standard deviation. It is especially effective in scenarios where the data within clusters may not follow a normal distribution, addressing the challenges posed by the diverse and potentially skewed data characteristics encountered in district heating systems.

To assess substation performance, we use the primary \(\Delta T\), often seen as a key performance indicator and reflects the efficiency of heat transfer of a substation. \(\Delta T\) is the difference between the primary supply temperature and the primary return temperature, with high \(\Delta T\) values suggesting better performance. However, since our analysis focuses on relative comparisons, a substation is considered poor-performing when its mean modified z-score is below 2$\sigma$ (single-sided) compared to its peers.

\subsection{Experimental Setup}
\label{sec:method:sub:experimental-setup}

We evaluate our approach against common time-series clustering methods in DH, as outlined in~\cite{van_dreven_2023}. Specifically, we use time series \(k\)-means, Spectral clustering, $k$-Shape, and SNN. To evaluate the compactness and connectivity of our clusters, we employ the mean MST intra-cluster distance metric, mean intra-cluster variance, and empirical observations. The mean MST intra-cluster distance is defined as:

\begin{equation}
    \text{MI} = \frac{1}{n} \sum_{i=1}^{n} d_i,
\end{equation}

\noindent where MI is the mean intra-cluster distance, \(d_i\) denotes the distance between connected nodes within the cluster, and \(n\) is the total number of nodes.

The mean intra-cluster variance is defined as:

\begin{equation}
    \text{MV} = \frac{1}{n} \sum_{i=1}^{n} (d_i - \overline{d})^2,
\end{equation}

\noindent where MV is the mean intra-cluster variance, \(d_i\) represents the distances between each pair of nodes within a cluster, and \(\overline{d}\) is the mean distance calculated from these \(d_i\) values. This metric quantifies the variability of distances within a fully connected subgraph of a cluster. A lower variance indicates a high consistency in node connectivity, suggesting that the substations are closely and uniformly integrated. In contrast, a high variance may highlight significant disparities in distances. While our proposed solution is applied to the supply temperature profiles, we validate our outcomes using the actual geographical substation locations. Such data is typically confidential and, therefore, not commonly accessible. However, their availability in this study allows us to rigorously assess our method's effectiveness in approximating the relative substation locations.

We compute the sensitivity and specificity of our anomaly detection model based on manual analysis provided by a domain expert. Sensitivity, or true positive rate, measures the proportion of actual positives the model correctly identifies. Specifically, it is defined as:

\begin{equation}
\text{sensitivity} = \frac{\text{true positives}}{\text{true positives} + \text{false negatives}}
\end{equation}

On the other hand, specificity, or the true negative rate, measures the proportion of actual negatives that are correctly identified and is given by:

\begin{equation}
\text{specificity} = \frac{\text{true negatives}}{\text{true negatives} + \text{false positives}}
\end{equation}

It is crucial to consider that the expert analysis, while invaluable, may have its own limitations and not encompass the full spectrum of anomalies. This potential oversight could influence the perceived accuracy of the sensitivity and specificity metrics.

\section{Results and Discussion}
\label{sec:results-discussion}

This section comprises two subsections, subsection~\ref{sec:results:sub:network} draw our findings on the relative topology approximation and identification of the supply temperature anomalies. At the same time, subsection~\ref{sec:results:sub:performance-anomalies} discusses anomaly detection performance and provides relative comparisons.

\begin{figure}[t]
    \centering
    \begin{subfigure}[b]{0.49\textwidth}
        \centering
        \includegraphics[width=\textwidth]{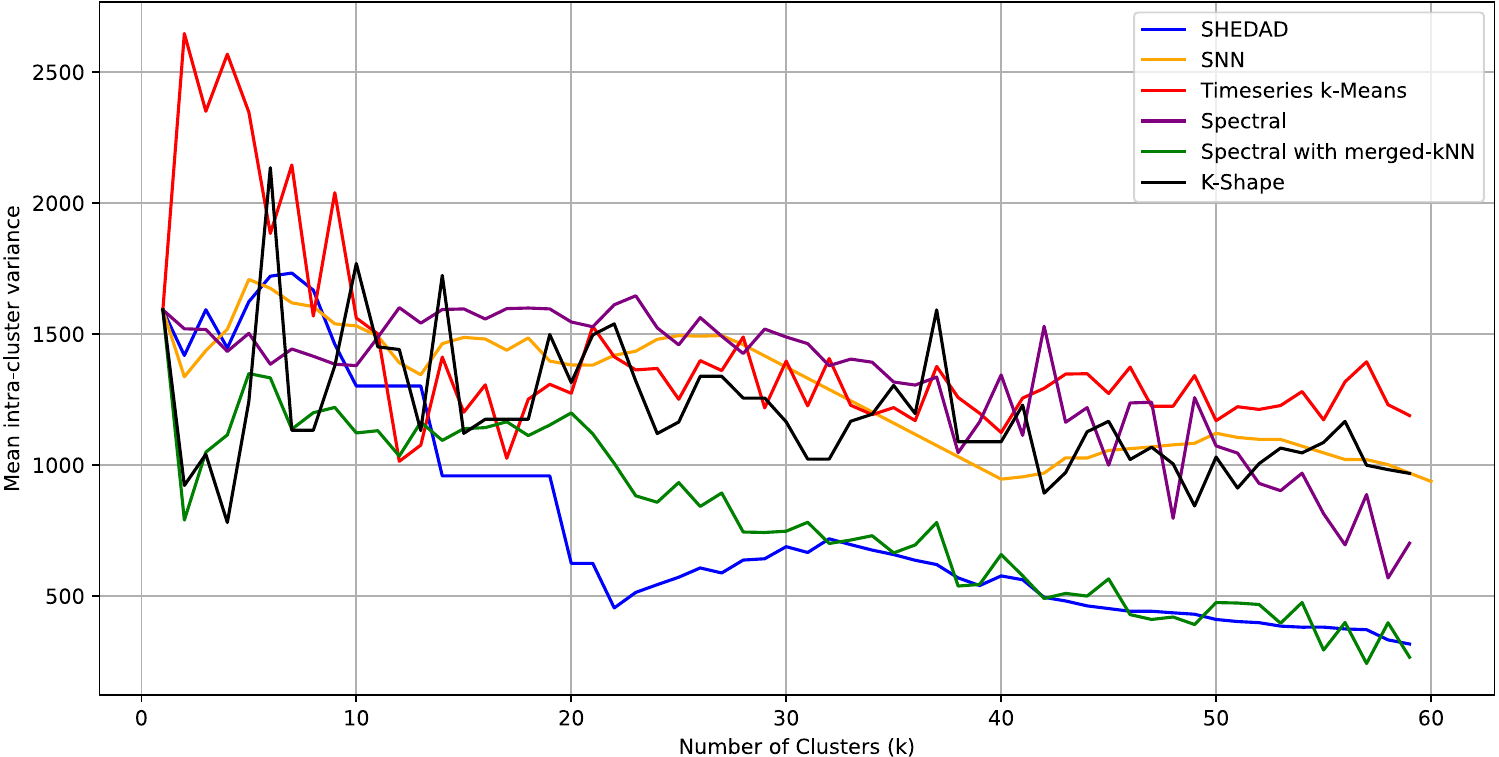}
        \caption{}
        \label{fig:evaluation:variance}
    \end{subfigure}
    \hfill
    \begin{subfigure}[b]{0.49\textwidth}
        \centering
        \includegraphics[width=\textwidth]{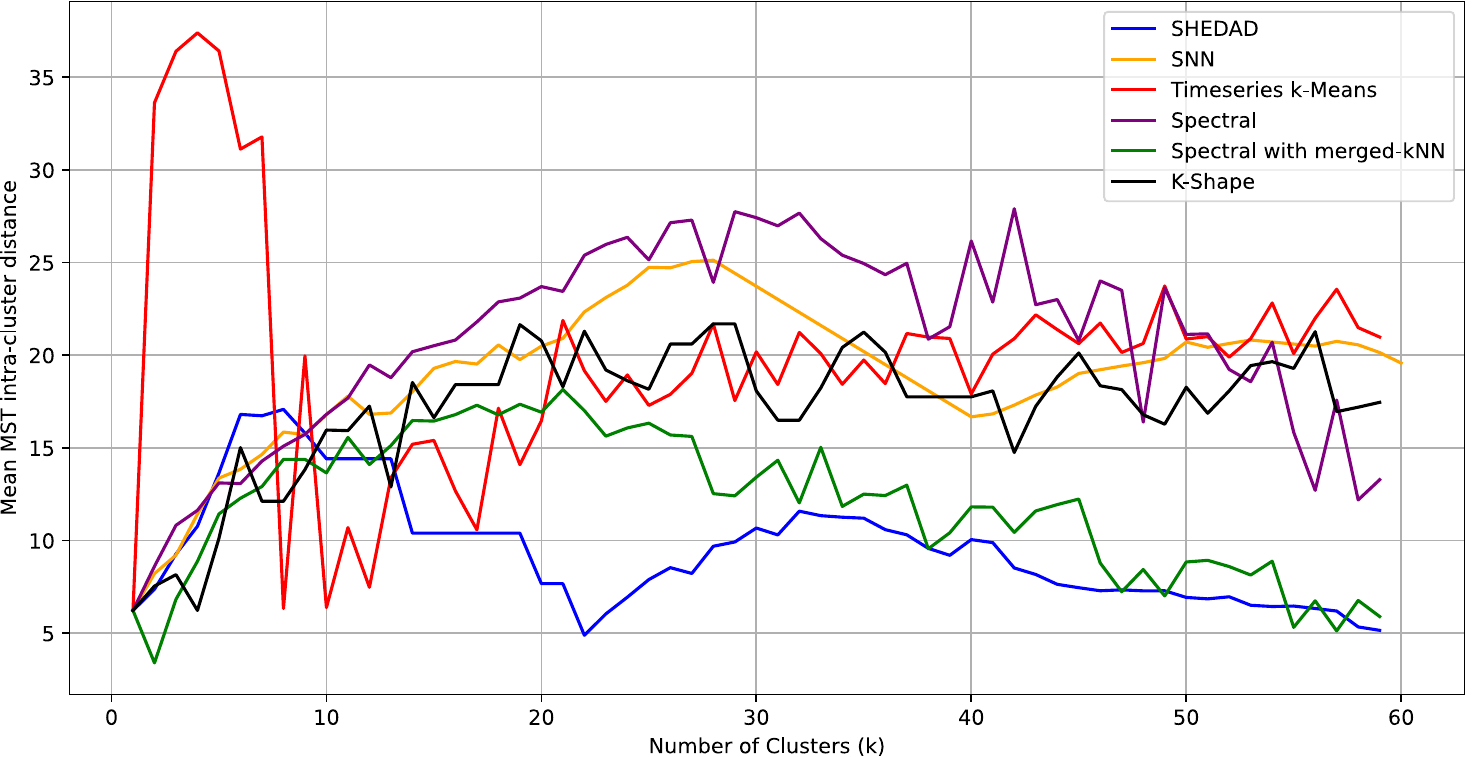}
        \caption{}
        \label{fig:evaluation:distance}
    \end{subfigure}
    \caption{Performance evaluation for various time-series clustering methods, with mean intra-cluster variance in (a) and mean DTW intra-cluster distance in (b).}
    \label{fig:evaluation}
\end{figure}

\subsection{Relative Topology Approximation}
\label{sec:results:sub:network}

Our study evaluated the performance of multiple clustering algorithms compared to our proposed method on DH data. The evaluation criteria included mean intra-cluster variance and MST distances. As shown in Figure\ref{fig:evaluation}, SHEDAD was consistently outperforming other clustering methods (from \(k > 13\)) and maintained a low mean intra-cluster variance (Figure~\ref{fig:evaluation:variance}) and distance (Figure~\ref{fig:evaluation:distance}) across various cluster sizes $k$. It is worth highlighting that the proposed approach can recognize both the degree of homogeneity and stability in cluster composition, indicating that the substations within each cluster are closely connected in proximity.

Specifically, the default SNN method demonstrated poor performance, consistently showing high mean intra-cluster variance and MST distances across various cluster sizes. This indicates its inability to capture densely connected clusters effectively. Additionally, it was not stable in finding outlying substations, which affected performance metrics.

$k$-Means displayed moderate performance, with low mean intra-cluster MST distance scores at \(k=8\) and \(k=10\). However, these scores were misleading, as visual validation indicated poorly formed uniform clusters, with one large cluster and many singletons resulting in a low score. The method exhibited considerable variability across different \(k\) and increased scores as the number of $k$ grew. Typically, the performance should increase, as seen in our method, as the number of $k$ grows. $k$-shape consistently showed a high mean intra-cluster variance and MST distances, indicating poor clustering formations. While $k$-shape may help to monitor network delays, as it focuses on the shape of data, these network delays do not translate well into a relative topology approximation; substations clusters showed consistent dispersion and non-uniformity. Noteworthy is spectral clustering. While initially, spectral clustering was among the least effective methods, a significant improvement was observed when applying our initialization process (adaptive k-NN graphs merged into a single graph). Notably, for larger values of k (greater than 30), our findings indicate that our enhanced spectral clustering and the proposed SHEDAD method exhibit congruent performance trends. The increase in performance indicates our initialization approach's broad applicability and effectiveness across different clustering algorithms.

\begin{figure}[!]
\centering
\includegraphics[width=\columnwidth]{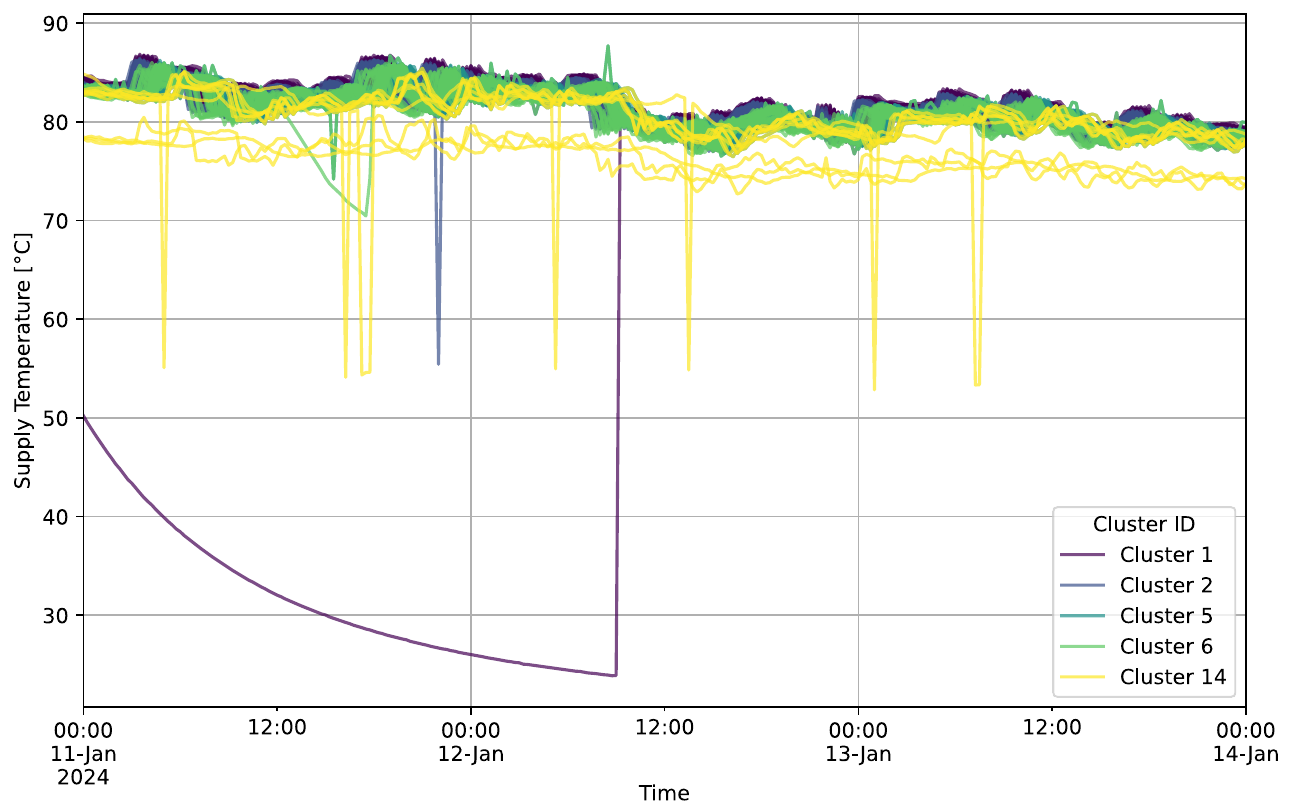}
\caption{Clusters of substations. Each line denotes the supply temperature profile of a substation, some of which include noise in the measurement data.}
\label{fig:cluster-noise}
\end{figure}

In Figure~\ref{fig:cluster-noise}, we present several clusters that highlight the impact of noise, which varies from sudden spikes in temperature readings (clusters 2, 5, 6, and 14) or short graduate loss of the supply temperature due to lack of heat demand (cluster 1). The SHEDAD method demonstrated a high degree of resistance to noise, effectively forming dense clusters despite the presence of disturbances, and highlights the capability of our approach to maintain cluster integrity under these noisy conditions.

\begin{figure*}[t]
    \centering
    \begin{subfigure}[b]{0.49\textwidth}
        \centering
        \includegraphics[width=\textwidth]{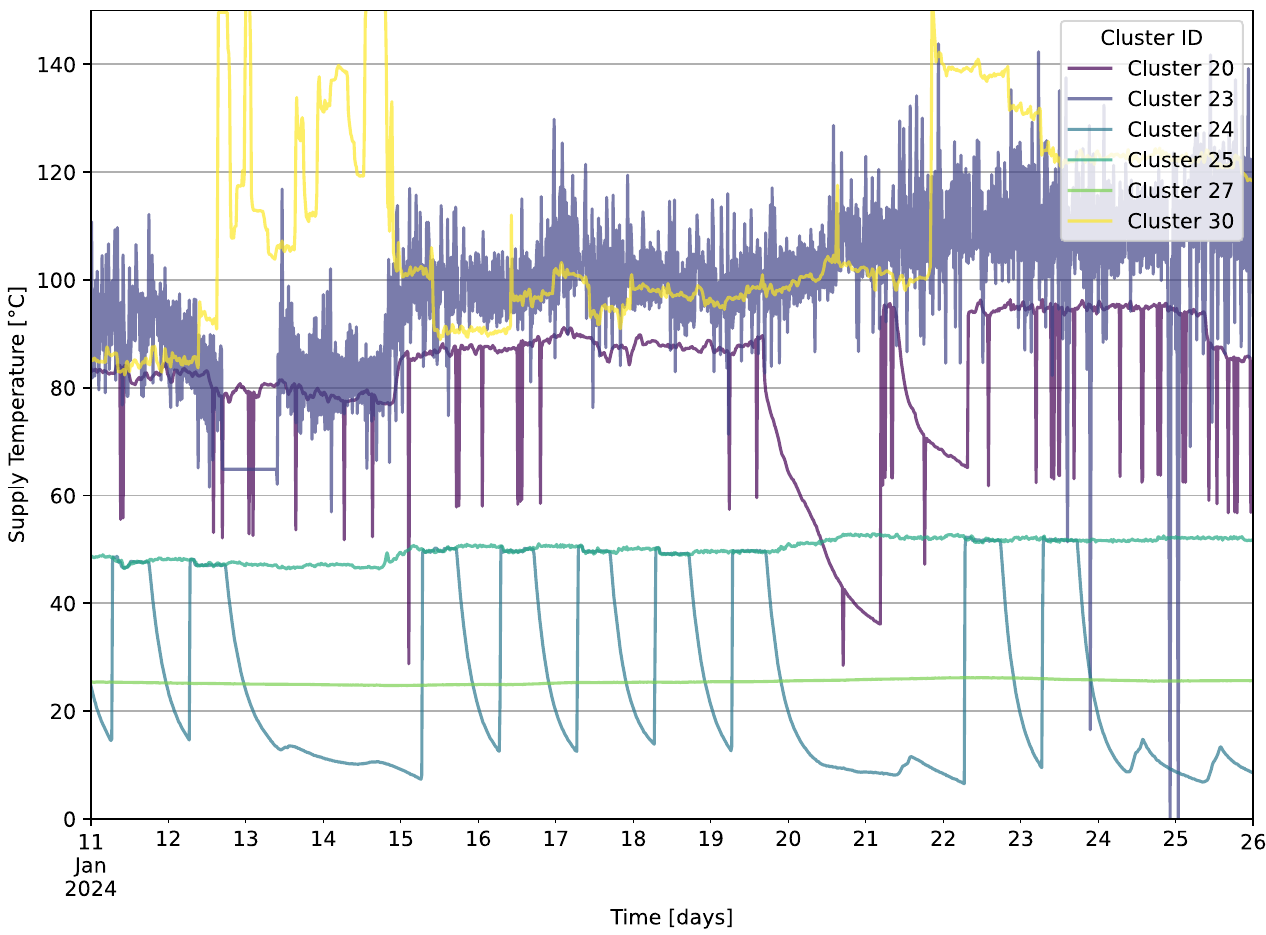}
        \caption{}
        \label{fig:cluster-outliers:supply}
    \end{subfigure}
    \hfill
    \begin{subfigure}[b]{0.49\textwidth}
        \centering
        \includegraphics[width=\textwidth]{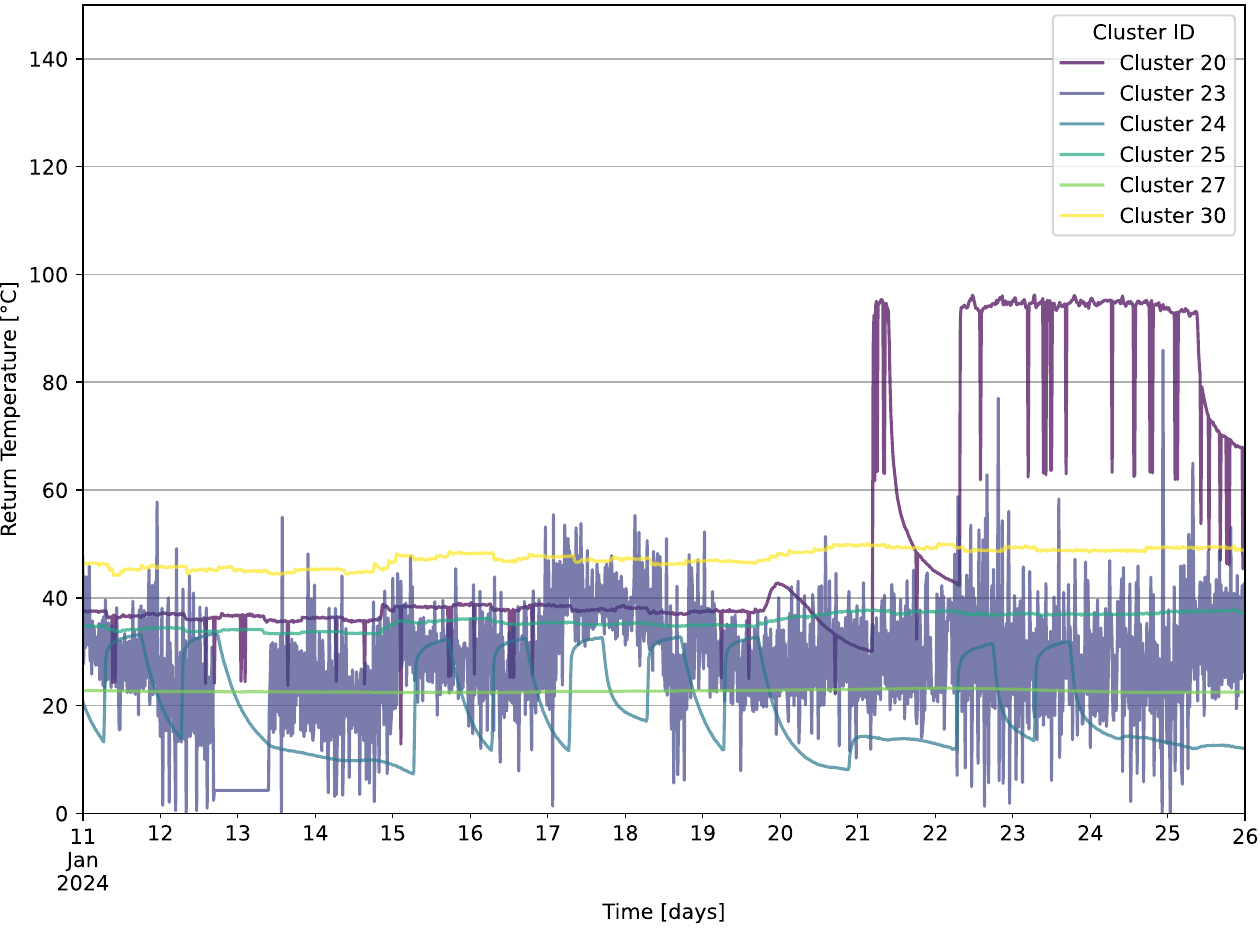}
        \caption{}
        \label{fig:cluster-outliers:return}
    \end{subfigure}
    \caption{Subset of supply temperature anomalies. Each substation is shown with its supply temperature (a) and return temperature (b), demonstrating substantial deviations and unusual patterns (cluster 20, 23, 27, and 30), while other substations (cluster 24 and 25) may be unusual though having normal operation.}
    \label{fig:cluster-outliers}
\end{figure*}

Additionally, we could effectively separate temperature anomalies into their own clusters. By dynamically adjusting the neighborhood size based on the density of connections (low-weight and high-weight thresholds), our approach separates these outlying components early in the initialization process. This adaptability is crucial for effectively isolating outlier substations that deviate from common patterns. These substations show substantial deviations from most operational patterns and thus are assigned a singleton cluster. The patterns vary from significant fluctuations, unusual behavior, missing readings, or sudden increases. Figure~\ref{fig:cluster-outliers} displays a subset of the total 16 supply temperature anomalies we found, showcasing only notable examples. Out of the total 16 supply temperature anomalies identified, ten were independently confirmed by a domain expert through manual analysis of the same dataset, underscoring the reliability of our method. 

For instance, cluster 24 displays unusual behavior compared to the large share of all substations and is assigned its own cluster. However, this deviation does not necessarily indicate a malfunction. The observed pattern of consistent heat consumption during the daytime with no consumption at night suggests this pattern may indicate the specific operational characteristics or building type associated with this substation. Similarly, Clusters 25 and 27 exhibited unusual behaviors. While consistent heat demand at low supply temperatures is not unusual, Cluster 27 exhibits an anomaly where the return temperature equals the supply temperature, resulting in a \(\Delta T\) of zero, which is highly unusual and signals further investigation. Other notable instances include Cluster 20, 23, and 30. Cluster 20 shows a higher return temperature than its supply temperature, which may indicate an issue in the system or be related to measurement errors or sensor malfunctions.
On the other hand, cluster 23 displays significant fluctuations in supply and return temperatures and may suggest sub-optimal performance, possibly due to an issue in the substation. Finally, cluster 30 shows a substantial increase in supply temperature, reaching up to 150{\textdegree C}, a condition not observed in other substations. While some substations may show unusual behavior but normal operational patterns, others show signs of more severe conditions. Thus, these supply temperature anomalies could form a basis for strategically focusing maintenance efforts. 

\subsection{Performance Anomalies}
\label{sec:results:sub:performance-anomalies}

The relative network approximation forms the basis for our performance anomaly detection. Initially, we identified 30 clusters, from which 16 clusters showing supply temperature anomalies were excluded, leaving us with 14 remaining clusters for the subsequent performance anomaly detection phase. As outlined in subsection~\ref{sec:method:sub:anomaly-detection}, we construct an MST for each of the 14 remaining clusters. These MSTs enable us to perform intra-cluster performance comparisons. Each substation is evaluated against its \(k\) (indirect) neighbors using modified z-scores. In each comparison, the respective substations with a mean modified z-score below 2$\sigma$, as compared to their neighboring substations, received an anomaly vote. Votes are normalized by the number of comparisons the substation was involved in, up to \(k-1\), resulting in a score between 0 (normal) and 1 (anomalous).

Through this process, we identified 14 substations that received at least one anomaly vote, indicating that compared to their peers, these substations are underperforming. Notably, five of these substations were consistently flagged across all their comparisons, each receiving the highest anomaly score of 1, which indicates a persistently poor performance relative to their peers and across all assessments. Two of these 14 anomalies were independently confirmed by a domain expert through manual analysis of the substations. Ideally, a normal operating substation should show a stable and consistent pattern in its return temperature or \(\Delta T\). Substantial variance or oscillating behavior may hint towards a fault in the substation, ranging from mechanical issues to wrong software settings. Upon further inspection, substations with an anomaly score of 1 exhibited this behavior, including high return temperatures, oscillating readings, low \(\Delta T\), or sudden temperature drops, potentially indicating poor heat utilization or other issues such as sensor malfunctions and incorrect settings. 

This approach allows for concentrating maintenance efforts on a smaller subset of poorly performing substations. By targeting these specific substations, network operators can enhance overall network performance and reliability and optimize the allocation of maintenance resources. Such an approach enables a cyclical or iterative approach to network maintenance, where utilities can initially focus on the most anomalous substations, e.g., an anomaly score of 1. Post-maintenance evaluations can re-assess the network, focusing on the next set of substations showing significant deviations. This iterative cycle ensures continuous improvement and stability in the operational performance of the DH network while providing an opportunity to collect labeled data on both normal operational behavior and fault instances, e.g., using a labeling taxonomy as outlined in~\cite{Mansson_Lundholm}. This collection of labeled data is crucial for advancing the field toward predictive maintenance capabilities instead of reactive fault management.

Our model achieved a sensitivity of approximately \(65\%\) and specificity of approximately \(97\%\). This score suggests potential areas for improvement in identifying anomalies but demonstrates a robust capacity to correctly identify non-anomalous substations and minimize false positives. This high level of specificity is critical as it helps to avoid unnecessary maintenance expeditions, thereby conserving resources and optimizing operational efficiency.

\subsection{Limitations}

While our methodology has demonstrated effectiveness in network topology approximation and anomaly detection of substations, some limitations should be considered. The performance of our method is sensitive to the choice of hyperparameters, such as the number of neighbors, warp distance, and edge inclusion/exclusion thresholds, which require specific domain knowledge that may not always be available. Additionally, at higher cluster counts, our approach tends to form a few large clusters and many singletons, potentially obscuring detailed insights and affecting the granularity of anomaly detection. 
Moreover, while the method's performance might vary across different DH networks, it is based on universal aspects of DH systems. Though validated only on one dataset, we expect the SHEDAD approach to be generalizable to other DH networks. Future research should test its adaptability and refine its performance across a diverse set of DH networks.

\section{Conclusions and Future Work}
\label{sec:conclusion-futurework}

This study introduced a novel SNN-enhanced approach (SHEDAD) for anomaly detection in DH networks, leveraging the relative network topology to identify operational anomalies. Our method adeptly manages the complex data structures of DH systems and maintains network confidentiality. It has effectively isolated anomalies, identifying 16 supply temperature anomalies and 14 performance anomalies among 248 substations. We achieved a sensitivity of approximately 65\% and a specificity of approximately 97\%. The results underscore the potential of our automatic method to improve the efficiency and reliability of DH networks by enabling more precise maintenance responses based on the identified anomalies. The proposed approach promotes a cyclical strategy for network maintenance, prioritizing intervention at substations with the highest anomaly scores. This iterative process enhances the operational performance of DH networks and facilitates the systematic collection of labeled data of both normal and faulty data. Such data are invaluable for training and validating data-driven models, improving anomaly detection and fault diagnosis capabilities over time. Possible future efforts are directed towards the following:

\begin{itemize}
\item Enhancing the method's network approximation capabilities, aiming to develop uniform-sized clusters and dynamically adjust the number \(k\) more effectively.
\item Extending validation and testing across various types of DH networks to improve adaptability and effectiveness under diverse conditions. We plan to integrate federated learning to enable decentralized, collaborative anomaly detection across substations without compromising data privacy.
\item Enhancing anomaly detection processes via machine learning techniques that harness federated learning for model training on data collected through targeted maintenance operations.
\end{itemize}

We aim to develop a more robust, efficient, and universally applicable anomaly detection methodology for DH networks by addressing these areas. 

\section*{Acknowledgement}
We acknowledge and thank \textit{Runa Smart Equipment} Co. Ltd. for providing the district heating data essential for this research.

\bibliographystyle{unsrtnat}

\bibliography{sample-base}
\end{document}